%% file: samplepaper.tex
\documentclass[runningheads]{llncs}
\usepackage[T1]{fontenc}
\usepackage{graphicx}
\usepackage{float}
\usepackage{booktabs}
\usepackage{mdframed}
\usepackage{subcaption}
\usepackage{multirow,soul,xcolor}
\usepackage{amsmath}
\usepackage{orcidlink}

\usepackage{tikz}
\usetikzlibrary{arrows.meta, positioning}
\usetikzlibrary{fit, positioning}
\usetikzlibrary{shapes.geometric} 
\usetikzlibrary{shapes.symbols}

\usepackage[most]{tcolorbox}

\tcbset {
  base/.style={
    arc=0mm, 
    bottomtitle=0.5mm,
    boxrule=0mm,
    colbacktitle=black!10!white, 
    coltitle=black, 
    fonttitle=\bfseries, 
    left=2.5mm,
    leftrule=1mm,
    right=3.5mm,
    title={#1},
    toptitle=0.75mm, 
  }
}

\definecolor{brandblue}{rgb}{0.34, 0.7, 1}
\newtcolorbox{mainbox}[1]{
  breakable,
  colframe=brandblue, 
  base={#1}
}

\newtcolorbox{subbox}[1]{
  colframe=black!30!white,
  base={#1}
}

\begin{document}
\title{Controlling the Mutation in Large Language Models for the Efficient Evolution of Algorithms}
\authorrunning{H. Yin et al.}
\titlerunning{Controlling Mutation in LLMs}
%
\author{Haoran Yin\inst{1}\orcidlink{0009-0005-7419-7488} \and Anna V. Kononova\inst{1}\orcidlink{0000-0002-4138-7024}
\and Thomas B{\"a}ck\inst{1}\orcidlink{0000-0001-6768-1478} \and Niki van Stein\inst{1}\orcidlink{0000-0002-0013-7969}}
\institute{LIACS, Leiden University, The Netherlands \\
\email{h.yin@liacs.leidenuniv.nl}
}
%
%
%
\maketitle
\begin{abstract}
The integration of Large Language Models (LLMs) with evolutionary computation (EC) has introduced a promising paradigm for automating the design of metaheuristic algorithms. However, existing frameworks, such as the Large Language Model Evolutionary Algorithm (LLaMEA), often lack precise control over mutation mechanisms, leading to inefficiencies in solution space exploration and potentially suboptimal convergence. This paper introduces a novel approach to mutation control within LLM-driven evolutionary frameworks, inspired by theory of genetic algorithms. Specifically, we propose dynamic mutation prompts that adaptively regulate mutation rates, leveraging a heavy-tailed power-law distribution to balance exploration and exploitation. Experiments using GPT-3.5-turbo and GPT-4o models demonstrate that GPT-3.5-turbo fails to adhere to the specific mutation instructions, while GPT-4o is able to adapt its mutation based on the prompt engineered dynamic prompts. Further experiments show that the introduction of these dynamic rates can improve the convergence speed and adaptability of LLaMEA, when using GPT-4o. This work sets the starting point for better controlled LLM-based mutations in code optimization tasks, paving the way for further advancements in automated metaheuristic design.
\end{abstract}

\section{Introduction}

The development of metaheuristic algorithms \cite{back1993overview} has been foundational for tackling complex optimization problems across domains such as engineering \cite{yang2013optimization}, artificial intelligence \cite{ojha2017metaheuristic}, and computational biology \cite{penas2015parallel}. Traditionally, these algorithms have relied on expert-crafted designs and manual tuning, a process that, while effective, is labor-intensive and constrained in scalability. Recent advances in Large Language Models (LLMs) \cite{chang2024survey,zhao2023survey} have brought new opportunities to this field, enabling automated generation and refinement of metaheuristics through natural language processing capabilities. The synergy between LLMs and evolutionary computation (EC) \cite{wu2024evolutionary} represents a transformative shift, leveraging LLMs for generating and enhancing code, while EC methods provide global search and iterative improvement capabilities.

A variety of methodologies exemplify this integration. The Large Language Model Evolutionary Algorithm (LLaMEA) \cite{van2024llamea} employs LLMs to iteratively generate, mutate, and select optimization algorithms, outperforming state-of-the-art methods in benchmark tasks. Similarly, the Algorithm Evolution using Large Language Models (AEL) framework and the Evolution of Heuristics approach \cite{liu2023algorithm,liu2024evolution} demonstrates how LLMs can autonomously evolve heuristics using in-context learning and feedback loops. Further, approaches such as Evolutionary Optimization with LLMs (EvoLLM) \cite{lange2024large} show how prompt engineering and mutation operators enhance LLMs’ ability to navigate search spaces effectively. These studies illustrate the potential for LLM-driven algorithmic frameworks to automate tasks traditionally requiring significant domain expertise.

Despite these advancements, existing frameworks often exhibit inefficiencies in exploring the solution space. A key limitation lies in the lack of precise control over mutation mechanisms during algorithm evolution, which can lead to redundant evaluations and suboptimal convergence. Addressing this gap, this paper introduces a novel mutation control mechanism for LLM-driven evolutionary frameworks. Drawing inspiration from genetic algorithms where a mutation rate sampled from a heavy-tailed power-law distribution is most beneficial \cite{doerr2017fast}, we propose dynamic mutation prompts that adaptively modulate the mutation rate to balance exploration and exploitation, aiming to enhance the convergence speed of LLaMEA.

\section{Related Work}
The generation and optimization of metaheuristic algorithms using LLMs has become a research hotspot in the field of research that explores the integration of large-scale language models with evolutionary computation, including EoH and LLaMEA~\cite{van2024llamea,liu2024evolution,wu2024evolutionary,van2024loop}. LLaMEA is the basis of our work as it provides an open-source modular framework that we can easily expend on. It is designed to utilize LLMs to automatically generate and improve optimization algorithms. The core idea is to achieve steady improvement of algorithm performance through an automated evolutionary process driven by language models, and an evolutionary strategy (1, 1) or (1 + 1), including algorithm generation, mutation and selection~\cite{beyer2002evolution}. Specifically, the LLaMEA process consists of the following steps:
\begin{itemize}
    \item Generation: The LLM models generate the algorithms based on the generation prompt. This prompt can include problem descriptions, constraints, and a framework of expected solutions. By adjusting the prompt, the model can be guided to generate more targeted algorithmic candidates.

    \item Mutation: When LLMs are tasked with generating a new algorithm, they receive not only the initial generation prompt, but also mutation prompts simultaneously. These mutation prompts guide the model to either make slight refinements to the existing algorithm or completely redesign it. The latter approach is analogue to a restart in Evolutionary Algorithm optimization, thereby enhancing the model's exploration capabilities.

    \item Selection: For the best current and newly generated algorithms, LLaMEA evaluates them based on performance metrics. Only the algorithm that excels in multidimensional metrics is allowed to proceed to the next iteration. This performance feedback-based selection mechanism enables LLaMEA to continuously optimize the quality of the generated algorithms.

\end{itemize}
Through this iterative process, LLaMEA enables the algorithms to evolve themselves, demonstrating greater efficiency and flexibility in solving complex problems~\cite{van2024llamea}. This framework not only reduces development time, but also generates algorithmic solutions with high performance on specific tasks, which has strong potential for practical applications. However, research in LLaMEA so far has mainly focused on the automatic generation of algorithms, while careful control of the mutation behavior during their evolution has been neglected, which makes the results of generation uncontrollable and leads to a potential waste of computational resources and money.

\begin{figure}[!t]
    \centering
    \includegraphics[width=0.8\linewidth]{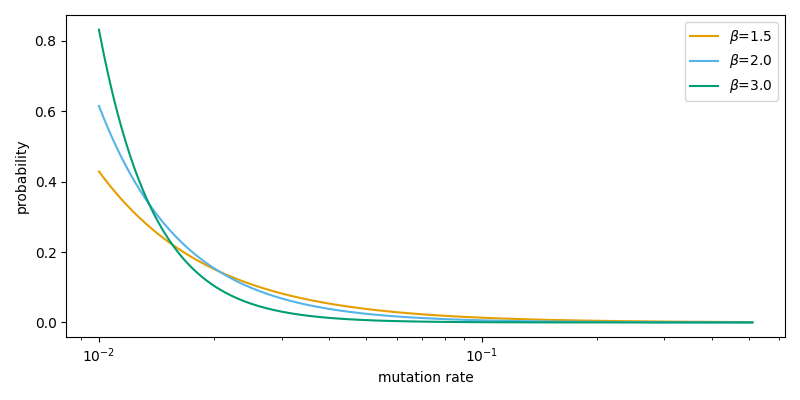}
    \caption{The distribution of Equation~\ref{eq:fmut}. The area under the curves are all 1.}
    \label{fig:fmut}
\end{figure} 

To explore the potential improvement of the mutation mechanism in LLaMEA, the choice of mutation rate is crucial. Traditional genetic algorithms recommend a fixed mutation rate, which is generally effective enough in simple unimodal optimization problems~\cite{holland1992adaptation,back1993overview}. However, for complex and multimodal optimization problems, dynamically adjusting the mutation rate has been shown to provide significant performance gains~\cite{jansen2006analysis,bottcher2010optimal}. Therefore, we refer to the fast mutation operator $fmut_{\beta}$ from B. Doerr et al.~\cite{doerr2017fast}, to help improve the convergence speed of LLaMEA.
$fmut_{\beta}$ is a heavy-tailed mutation operator for mutation steps, which is specifically beneficial for dealing with multimodal functions, where static mutation rates may not be efficient~\cite{hansen2001completely,antipov2020runtime}. It employs a randomized mutation rate $\alpha / n$, where $\alpha$ is an integer that follows a power-law distribution that does not concentrate strongly around its mean, and $n$ represents the length of the bit string. This allows for a greater range of mutation strengths, potentially improving the algorithm's ability to escape local optima and explore the fitness landscape more effectively. This power-law probability distribution is represented as Equation~\ref{eq:fmut}, and its distribution is illustrated in Figure~\ref{fig:fmut}:
\begin{equation}
    Pr(\alpha/n) = \left ( \sum_{i=1}^{\lfloor n/2 \rfloor} i^{-\beta} \right )^{-1} \alpha^{-\beta}
\label{eq:fmut}
\end{equation}
Studies show that using a heavy-tailed mutation rate significantly reduces the optimization time for complex problems compared to a fixed mutation rate, and this approach has been proven to optimize functions more efficiently, particularly for complex problems with multiple local optima~\cite{doerr2017fast}.

\section{LLaMEA with Dynamic Mutation Rate} \label{sec:mutationrate}
By integrating dynamic mutation rates from discrete
power-law distributions into the LLaMEA framework, we can simulate the mutation process in genetic algorithms with a view to improving the performance and adaptability of the algorithms to complex and multimodal optimization problems.

The effort of introducing a dynamic mutation rate to LLaMEA starts with the refinement of the existing mutation prompt. In the LLaMEA framework, each generated algorithm is produced by a LLM using the parent algorithm and a mutation prompt, but this mutation prompt is simply 'Either refine or redesign to improve the algorithm'. To make this prompt more specific, we modified it to include the information about the mutation rate we want. The question of \textit{whether LLMs can correctly understand and accurately execute the mutation prompt} needs to be investigated. Once this is established, mutation rate should be probabilistic and follow a given distribution, inspired by $fmut_{\beta}$.


We start with experiments to establish whether LLMs can achieve a fixed mutation rate when given a mutation prompt. We measure the code mutation effect by: first computing the logarithm of the ratio of the percentage of code modification of the actual mutated code (child) to that of the initial code (parent) to the requested mutation rate and then calculating the mean squared error (MSE) as shown in Equation~\ref{eq:mse}:
\begin{equation}
    MSE(x_i) = \frac{\sum_{j=2}^{N}\left ( \log{\frac{x_i^j}{x_i}} - \log{\frac{x_i}{x_i}} \right )^2}{N-1} = \frac{\sum_{j=2}^{N}\log^2{\frac{x_i^j}{x_i}}}{N-1}
\label{eq:mse}
\end{equation}
where $x_i$ is requested mutation rate and $x_{i}^j$ is delivered code difference by LLaMEA of $j$-th algorithm code, $N$ is the number of generated algorithm codes. To account for the distribution of mutation rates, we define target-distribution-weighted score (TDW-score) to measure the overall effect of different prompts as shown in Equation 3:
\begin{equation}
    \text{TDW-score}(prompt) = \frac{\sum_{i=1}^{M}w_i MSE(x_i)}{M}
\label{eq:score}
\end{equation}
where $x_i$ is $i$-th mutation rate we experiment with, $w_i=\frac{Pr(x_i)}{\sum_{j=1}^{M}Pr(x_j)}$ is the weight of $MSE(x_i)$ and $M$ is the number of experimented mutation rates. With the MSE and the TDW-score, we can visualize the difference between the effects of different investigated prompts more intuitively.

We can then explore the \textit{impact of mutation prompts on the LLaMEA generation algorithm when combined with the dynamic mutation rate}. We set the number of lines of the parent code to $n$ in Equation~\ref{eq:fmut}, and the mutation rate obtained by sampling from Equation~\ref{eq:fmut} at each mutation operation replaces $x$ in the mutation prompt, and we use the same performance metric as LLaMEA~\cite{van2024llamea}, \textit{Area Over the Convergence Curve} (AOCC), as introduced in~\cite{van2024explainable}.

\subsection{Manually Constructed Prompts}
To explore how to make mutation prompts effective, we start with the most basic and intuitive \textbf{prompt 1}, which contains only a description of the mutation task and a indication of the mutation rate. Then, step by step, we continue to add new content, including mandatory clarification (\textbf{prompt 2}), strict limitations (\textbf{prompt 3}), example specification (\textbf{prompt 4}), and context-specific requirements (\textbf{prompt 5}). The prompts are shown below: 

\begin{mainbox}{Manual prompts}
    \begin{description}
        \item[\textbf{Prompt 1}] Now, refine the strategy of the selected solution to improve it. Make sure that you only change $x\%$ of the code.
        \item[\textbf{Prompt 2}] Now, refine the strategy of the selected solution to improve it. Make sure that you only change $x\%$ of the code. This changing rate $x\%$ is the mandatory requirement.
        \item[\textbf{Prompt 3}] Now, refine the strategy of the selected solution to improve it. Make sure that you only change $x\%$ of the code. This changing rate $x\%$ is the mandatory requirement, you cannot change more or less than this rate.
        \item[\textbf{Prompt 4}] Now, refine the strategy of the selected solution to improve it. Make sure that you only change $x\%$ of the code, which means if the code has 100 lines, you can only change $x$ lines, and the rest lines should remain the same. This changing rate $x\%$ is the mandatory requirement, you cannot change more or less than this rate.
        \item[\textbf{Prompt 5}] Now, refine the strategy of the selected solution to improve it. Make sure that you only change $x\%$ of the code, which means if the code has 100 lines, you can only change $x$ lines, and the rest lines should remain the same. For this code, it has $n$ lines, so you can only change $\lfloor n \times x / 100 \rfloor$ lines, the rest $n - \lfloor n \times x / 100 \rfloor$ lines should remain the same. This changing rate $x\%$ is the mandatory requirement, you cannot change more or less than this rate.
    \end{description}
\end{mainbox}

\subsection{Automatic Generated Prompts}
\label{sec:methodology:Automatic Generated Prompts}
Popular research at the forefront has shown that LLMs themselves have capabilities comparable to human prompt engineers~\cite{zhou2022large}. Therefore, we also briefly explore initially the effects of LLM-generated mutation prompts. We passed the following meta-prompt to ChatGPT 4~\cite{achiam2023gpt}:

\begin{subbox}{Meta prompt} 
Now, imagine yourself as a prompt engineer, you give <LLM> a piece of optimization algorithm code, and you want <LLM> to modify it by $x$\% (where $x$ is a predefined number from 2 to 40, and indicated the code difference between the new one and the old one) exactly to improve the algorithm performance (meaning optimization convergence speed, not code efficiency, for example, by trying different mutation, selection, etc. strategies), what prompt would you give? Please give me at least 3 examples. Do not propose any specific directions or elements to change, since we want to cover the whole algorithm search space.
\end{subbox}
where <LLM> should be replaced with the specific model name when using this meta-prompt.

\section{Experimental Setup}

\noindent \textbf{Large Language Models}\\
This study uses two different large language models, GPT-4o and GPT-3.5-turbo, which have advanced capabilities and performance in generating meta-heuristic optimization algorithms.
\begin{itemize}
    \item \textbf{GPT-4o-2024-08-06}: this is OpenAI's latest large-scale multi-modal language model, known for its powerful ability to understand and generate complex code.
    \item \textbf{GPT-3.5-turbo-0125}: an earlier version of the GPT model, which lags behind the newer models but still shows good performance when dealing with complex tasks.
\end{itemize}

We choose these two models due to their popularity and ease of implementation. In future work it would be interesting to extend the experimental setup with additional (open-weight) models.

\medskip

\noindent \textbf{Choice of Fixed Mutation Rate}\\
Since we expect to achieve dynamic mutation rates sampled from a heavy-tailed power-law distribution, and with a maximum mutation rate of no more than 50\% of the distribution, we choose 2\%, 5\%, 10\%, 20\%, and 40\% as important references we use to determine how well the LLM is able to mutate the code according to these rates.

\medskip

\noindent \textbf{Automatically Generated Prompts}\\
Using the meta-prompt of Section~\ref{sec:methodology:Automatic Generated Prompts}, we obtain \textbf{prompt 6} to \textbf{prompt 11}, where \textbf{prompt 6} to \textbf{prompt 8} correspond to the model GPT-3.5-turbo and \textbf{prompt 9} to \textbf{prompt 11} correspond to the GPT-4o model. These prompts are shown below:
\begin{mainbox}{Generated prompts for GPT-3.5-turbo} \label{box:gpt35}
\begin{description}
    \item[\textbf{Prompt 6}] Adjust the code such that the algorithm's convergence speed is improved, while ensuring that the changes result in an exact difference of $x\%$ compared to the original code. This difference should reflect the modification in functionality, not code style or syntax. Feel free to adjust any part of the algorithm (e.g., initialization, selection, mutation, or other components) to achieve faster convergence while maintaining the specified code difference.
    \item[\textbf{Prompt 7}] Modify the optimization algorithm code to improve its performance in terms of convergence speed. The modification should result in a code difference of exactly $x\%$. Ensure that the changes are meaningful to enhance optimization speed without focusing on code efficiency or readability improvements. Explore any strategy within the algorithm to achieve this, but keep the difference precisely at the specified percentage.
    \item[\textbf{Prompt 8}] Please enhance the convergence speed of the optimization algorithm given below by modifying it. The modifications should introduce a code difference of precisely $x\%$ compared to the original code. Focus on optimizing the algorithm's behavior rather than its implementation efficiency. You are free to explore any area of the algorithm's logic, but ensure that the total code difference remains exactly at $x\%$ and is geared toward faster convergence.
\end{description}
\end{mainbox}

\begin{mainbox}{Generated prompts for GPT-4o} \label{box:gpt4o}
\begin{description}
    \item[\textbf{Prompt 9}] Here's a piece of code for an optimization algorithm. Please modify it by exactly $x\%$ to improve the algorithm's performance in terms of optimization convergence speed. Focus on introducing meaningful changes that can potentially enhance its effectiveness, such as exploring alternative strategies or approaches across any aspect of the algorithm. Keep the modifications strictly within the specified $x\%$ range for code difference while striving for faster convergence.
    \item[\textbf{Prompt 10}] Take this code of an optimization algorithm and adjust it by $x\%$ to improve convergence speed. Make sure the modifications cover a broad spectrum of possible algorithm adjustments, considering changes across different components without exceeding $x\%$ in code difference. Your changes should aim to improve the algorithm's ability to reach optimal solutions more quickly.
    \item[\textbf{Prompt 11}] Please transform this optimization algorithm code by exactly $x\%$ in a way that enhances convergence speed. Keep the code difference precisely at $x\%$, and focus solely on achieving performance improvements through algorithmic adjustments across various elements of the code. Avoid focusing on code efficiency; instead, prioritize exploration of diverse approaches within the allowed modification percentage.
\end{description}
\end{mainbox}

\subsection{Experimental design}
\input{ES_tikz}

In our experiments, LLaMEA will be applied to the following scenario: using an evolutionary strategy (1 + 1) to propose and improve metaheuristic algorithms to solve single-objective continuous optimization problems, which is shown in Fig.~\ref{fig:es}. In each iteration, LLaMEA mutates the parent metaheuristic algorithm to obtain a child algorithm, and only the better algorithm of them can be the new parent for the next iteration. To determine which algorithm is better, the parent and child will be tested for their effectiveness with the widely used Black-Box Optimization Benchmarking (BBOB) test set~\cite{hansen2009real}, as well as with the IOHexperimenter platform~\cite{IOHexperimenter}, with mean AOCC as the measure of algorithm performance~\cite{van2024explainable}. The dimensionality is set to 5 for all BBOB problems, the tests cover 3 instances of each problem, and the generated metaheuristic algorithms are repeated three times to eliminate experimental randomness and ensure reliable results.

The experiments consisted of two parts: First, we explore the reliability of the mutation prompts. The second experiment aims to explore the possible enhancements that the dynamic mutation rate brings to LLaMEA. The following is the setup of the first experiment:
\begin{description}
    \item[Models] GPT-3.5-turbo-0125 and GPT-4o-2024-08-06 
    \item[Evolutionary strategy] (1 + 1)
    \item[Requested mutation rates] 2\%, 5\%, 10\%, 20\% and 40\%
    \item[Prompts used] a total of 11 different prompts, of which 5 are manually constructed to be generic and automatically generated to be model-specific, i.e., both GPT-3.5-turbo and GPT-4o are experimented with 8 prompts.
    \item[Number of independent runs] the experiment is repeated 3 times for each combination (model, requested mutation rate, prompt).
    \item[Code generation budget] 100 code instances are generated for each configuration.
\end{description}
After finishing the first experiment, we select the best-performing manually constructed prompts and automatically generated prompts for GPT-3.5-turbo and GPT-4o, respectively, and then experiment the effect of using a dynamic mutation rate on the basis of these prompts in LLaMEA. The setup for this followup experiment is as follows:

\begin{description}
    \item[Models] GPT-3.5-turbo-0125 and GPT-4o-2024-08-06 
    \item[Evolutionary strategy] (1 + 1)
    \item[Requested mutation rates] dynamic mutation rate, with $\beta = 1.5$ in Equation~\ref{eq:fmut}
    \item[Prompts used] the best-performing manually constructed prompts and automatically generated prompts for GPT-3.5-turbo and GPT-4o, respectively
    \item[Number of independent runs] the experiment is repeated 5 times for each combination (model, prompts).
    \item[Code generation budget] 100 code instances are generated for each configuration.
\end{description}

Through this exhaustive experimental setup, we aim to comprehensively assess the role of dynamic mutation in automated algorithm generation and its benefits.

\section{Results}

\subsection{Reliability of Mutation Prompts}
In Fig.~\ref{fig:score}, the MSE and TDW-scores are shown. $x$-axis and $y$-axis represent requested mutation rates and prompts, respectively. For both Figs.~\ref{fig:score-gpt-3.5-turbo} and \ref{fig:score-gpt-4o}, \input{score}the first five columns from the left illustrate MSE values, and the last column on the right shows the TDW-score, which is the most important metric for selecting the prompts. The smaller the value, the better the prompt. Thus, we can find that prompt2 is the best manually constructed prompt and prompt7 is the best automatic generated prompt for GPT-3.5-turbo, but for GPT-4o, prompt5 and prompt9 are the best of manually constructed prompts and best automatic generated prompts, respectively.

In addition, there is a very noticeable difference for both models: the MSE and TDW-scores of GPT-3.5-turbo are significantly inferior to those of GPT-4o. In order to explore the details behind these numbers in depth, we show in detail the distribution of code difference of the codes generated with different prompts and requested mutation rates, as well as a scatterplot of the ratio of actual delivered code difference to requested mutation rate for LLaMEA, displayed in Fig.~\ref{fig:aggregated}.\input{aggregated_plots}

In Figs.~\ref{fig:aggregated-gpt-3.5-turbo} and \ref{fig:aggregated-gpt-4o}, the red dashed line represents the requested mutation rate, and the green dotted-dashed line indicates that the delivered code difference of the points on this line is equal to the requested mutation rate. Each column of data contains mutated codes from the 100 codes generated in 3 independent runs using LLaMEA and the requested mutation rate in each prompt. 

From Fig.~\ref{fig:aggregated-gpt-3.5-turbo}, we can see that progressively more complex and accurate artificially constructed prompts do not improve the performance of LLaMEA with GPT-3.5-turbo. Moreover, it seems that GPT-3.5-turbo does not understand the mutation rate variation well. The code difference of the code generated by GPT-3.5-turbo always stays high regardless of whether the requested mutation rate is high or low, and the distribution is not concentrated.

From Fig.~\ref{fig:aggregated-gpt-4o}, we find that the distribution of code difference for GPT-4o is more concentrated compared to the result of GPT-3.5-turbo, and also closer to the requested mutation rates, suggesting that GPT-4o is more powerful. Therefore, GPT-4o is a significantly better choice than GPT-3.5-turbo for LLaMEA when economic conditions allow. Second, as the artificially constructed prompts become more complex (longer), the delivered code difference gets closer to the mutation rate, that is, the scatterplot gets closer and closer to the green dotted dashed line, especially when the mutation rate is 2\%, 5\%, and 10\%. This is encouraging, and it shows that GPT-4o understands the task of mutation prompts, and that prompt engineering is effective for mutation prompts and GPT-4o performs the mutation task quite well.

\subsection{Influence of Dynamic Mutation Rate}
\input{convergence} 

Fig.~\ref{fig:convergence} compares the convergence speed of the baselines and LLaMEA with the dynamic mutation rate as introduced in Section \ref{sec:mutationrate} using the best mutation prompt as identified before. Baselines are raw data directly from~\cite{van2024llamea}, which represent the default setting of LLaMEA without dynamic mutation rate. Shaded areas denote the standard deviation of the best-so-far. 

From Fig.~\ref{fig:convergence}, we can observe that the manually constructed mutation prompt with dynamic mutation rate using GPT-4o improves the performance of LLaMEA with the (1 + 1) evolutionary strategy, but on the other hand, the automatically generated mutation prompt is not much different from the baseline (even slightly worse). Considering that this experiment is relatively simple automatic prompt engineering, there is still room for exploring more efficient automatic prompt generation methods to define better mutation prompts.

For the GPT-3.5-turbo model, the prompt used seems to have little effect on the convergence speed, and both are weaker than the baseline (though not significantly). The overall performance of GPT-3.5-turbo is also significantly weaker than that of GPT-4o, which implies that the comprehension capacity of GPT-3.5-turbo becomes a bottleneck in improving the effectiveness of LLaMEA.

\input{code_diff}
Figure~\ref{fig:code_diff} shows the change in the code difference of the code generated by LLaMEA after applying the prompts of the dynamic mutation rate. Dynamically required mutation rates are generally small, and compared to GPT-3.5-turbo, GPT-4o is significantly easier to mutate at the required mutation rate. Without specifying a specific value for the mutation rate, both GPT-3.5-turbo and GPT-4o are more likely to carry out extreme mutations, as shown in Figs.~\ref{fig:code_diff-gpt-3.5-turbo-baseline} and \ref{fig:code_diff-gpt-4o-baseline}.

\section{Discussions and Future Work}
In this study, we observe that the design of mutation prompts has a significant impact on the ratio of code mutations applied by the LLM in an algorithm optimization task, and additionally influences the optimization performance. The experimental results show that there is a difference in performance between the manually constructed prompts and the automatically generated prompts, suggesting that the precise design of the prompts is crucial to the optimization effect of the algorithm. For the GPT-3.5-turbo model, the performance is not satisfactory even when the indication of the mutation rate is more explicit, which may be related to the model's ability to understand the mutation task.

The introduction of dynamic mutation rates can improve the adaptability of the LLaMEA algorithm and lead to better generated meta-heuristic algorithms. By experimental comparison, the LLaMEA framework using dynamic mutation rate has better convergence speed and optimization results than the default prompt without mutation rate when using GPT-4o. This finding supports the effectiveness of using dynamic control strategies in evolutionary algorithms. Despite the experimental success, there are still challenges in precisely controlling the mutation rate. In addition, advanced automatic prompt engineering techniques may require more development and testing in practice to produce and optimize more efficient algorithms.

Future research should consider developing more advanced automated cue engineering techniques. Research should also be extended to a wider variety and newer versions of local LLMs to assess the scalability and adaptability of mutation strategies and reduce funding consumption.

%
%
%
\bibliographystyle{splncs04}
\bibliography{citiations}
\end{document}

%% file: ES_tikz.tex
\begin{figure}[!tb]
    \centering
    \resizebox{\textwidth}{!}{  
    \begin{tikzpicture}[node distance=1cm and 1cm, box/.style={draw, rounded corners, align=center, minimum height=0.5cm, minimum width=2cm}, arrow/.style={->, thick}]
        \node[box, draw, align=center] (start) {generate initial\\code instance};
        \node[box, draw, align=center, right=0.7cm of start] (mutation) {prompt LLM to mutate\\ current code instance with\\a \textit{requested} mutation rate,\\compute code difference\\\textit{delivered} by LLM};
        \node[box, draw, align=center, above=0.9cm of mutation, outer sep=0pt] (eval) {evaluate code instance using\\IOHexperimenter with average AOCC\\ as performance measure on full BBOB\\suite in specified dimentionality};
        \node[box, draw, align=center, right=0.8cm of mutation] (replace) {declare mutated\\code instance\\as current};
        \node[diamond, draw, align=center, right=0.6cm of eval, aspect=2] (selection) {is mutated\\better than\\ parent?};
        \node[diamond, draw, align=center, right=0.8cm of replace, aspect=2] (end) {is budget\\exhausted?};
        \node[box, draw, align=center, right=0.9cm of end] (done) {return\\current\\code\\instance};
        
        \draw[->] (start) -- (mutation);
        \draw[<->] (start) |- (eval);
        \draw[->] (mutation) -- (eval);
        \draw[->] (eval.east) -- (selection.west);
        \draw[->] (selection.south) -| (replace.north) node[midway, left] {Yes};
        \draw[->] (selection.east) -- (end.north) node[midway, right] {No};
        \draw[->] (replace.east) -- (end.west);
        \draw[->] (end.east) -- (done.west) node[midway, above] {Yes};
        \draw[->] (end.south) -- ++(0,-0.9) node[midway, right] {No}  -- ++(-7.845,0) -- ++(0,0.7) -- (mutation.south);

    \end{tikzpicture}
    }
    \caption{LLaMEA with controlled mutation via dynamic prompts.} \label{fig:es}
\end{figure}
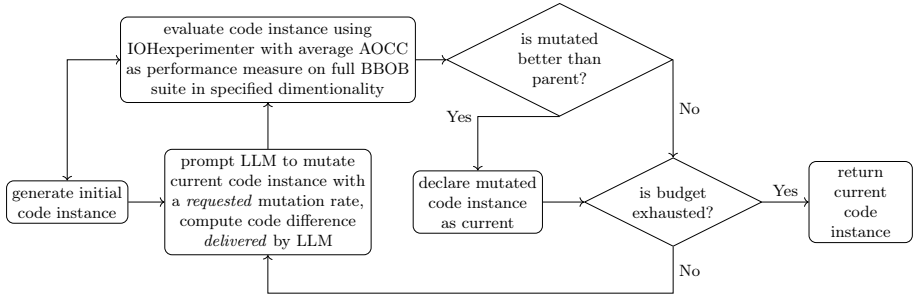

%% file: score.tex
\begin{figure}[!b]
  \centering
  \begin{subfigure}[b]{0.45\textwidth}
    \includegraphics[width=\textwidth]{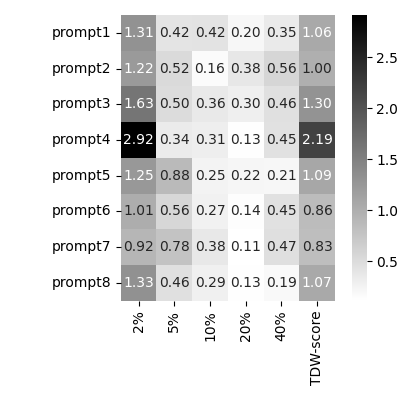}
    \caption{GPT-3.5-turbo}
    \label{fig:score-gpt-3.5-turbo}
  \end{subfigure}
  \hfill
  \begin{subfigure}[b]{0.45\textwidth}
    \includegraphics[width=\textwidth]{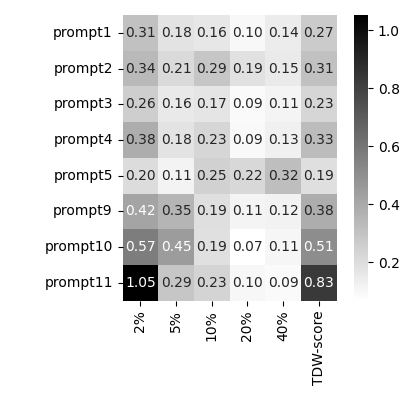}
    \caption{GPT-4o}
    \label{fig:score-gpt-4o}
  \end{subfigure}
  \caption{
  MSE and TDW-scores of prompts. For both figures, the $x$-axis is the requested mutation rates, the $y$-axis is the prompts, the first five columns from left show the corresponding MSE of different prompts with requested mutation rate, and the last column on the right shows the TDW-score of prompts. The smaller the value, the better the prompt.
  }
  \label{fig:score}
\end{figure}

%% file: aggregated_plots.tex
\begin{figure}[htp]
  \centering
  \begin{subfigure}[b]{\textwidth} 
    \includegraphics[ width=\textwidth]{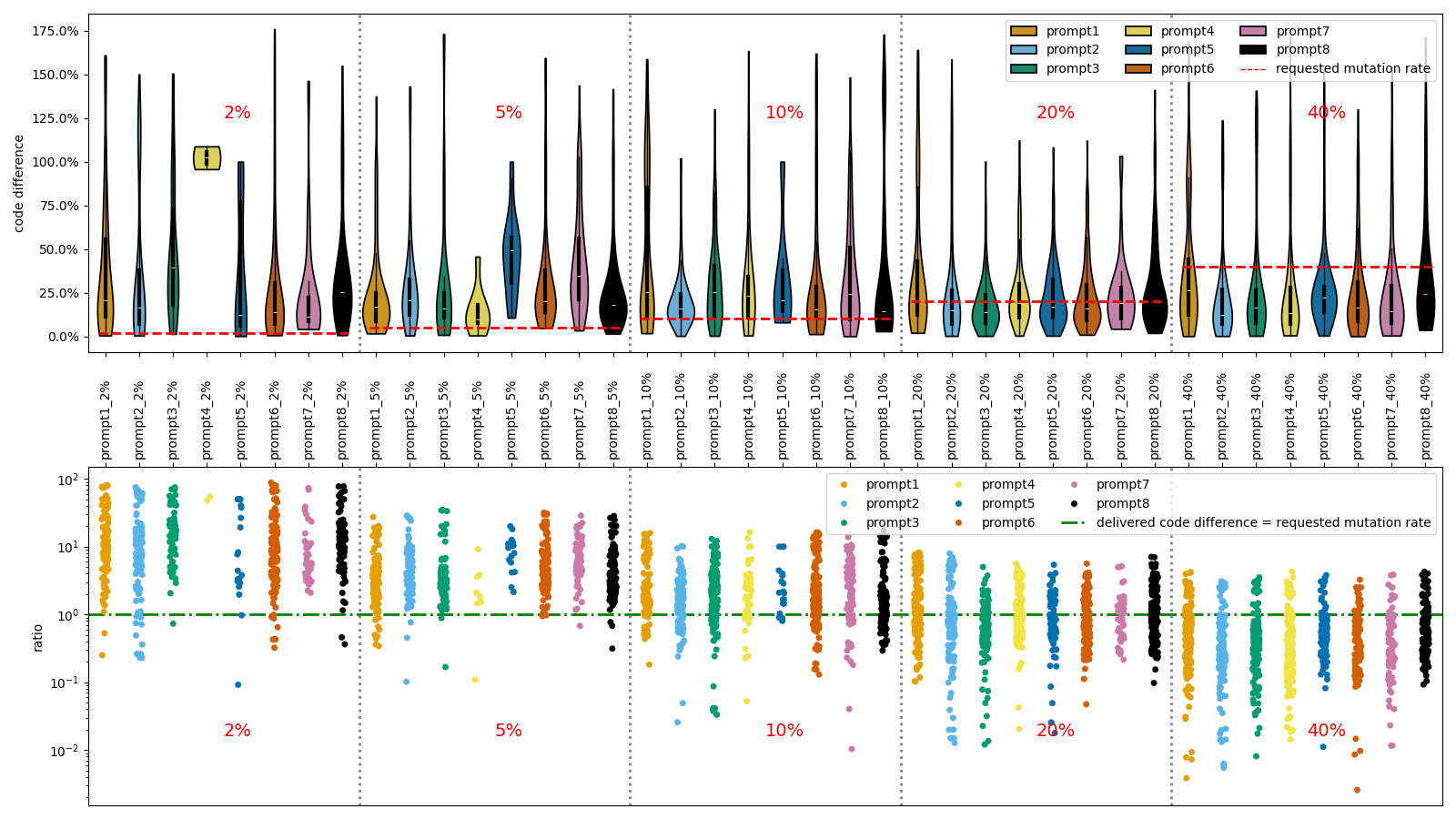} 
    \caption{GPT-3.5-turbo}
    \label{fig:aggregated-gpt-3.5-turbo}
  \end{subfigure}

  \begin{subfigure}[b]{\textwidth}
    \includegraphics[width=\textwidth]{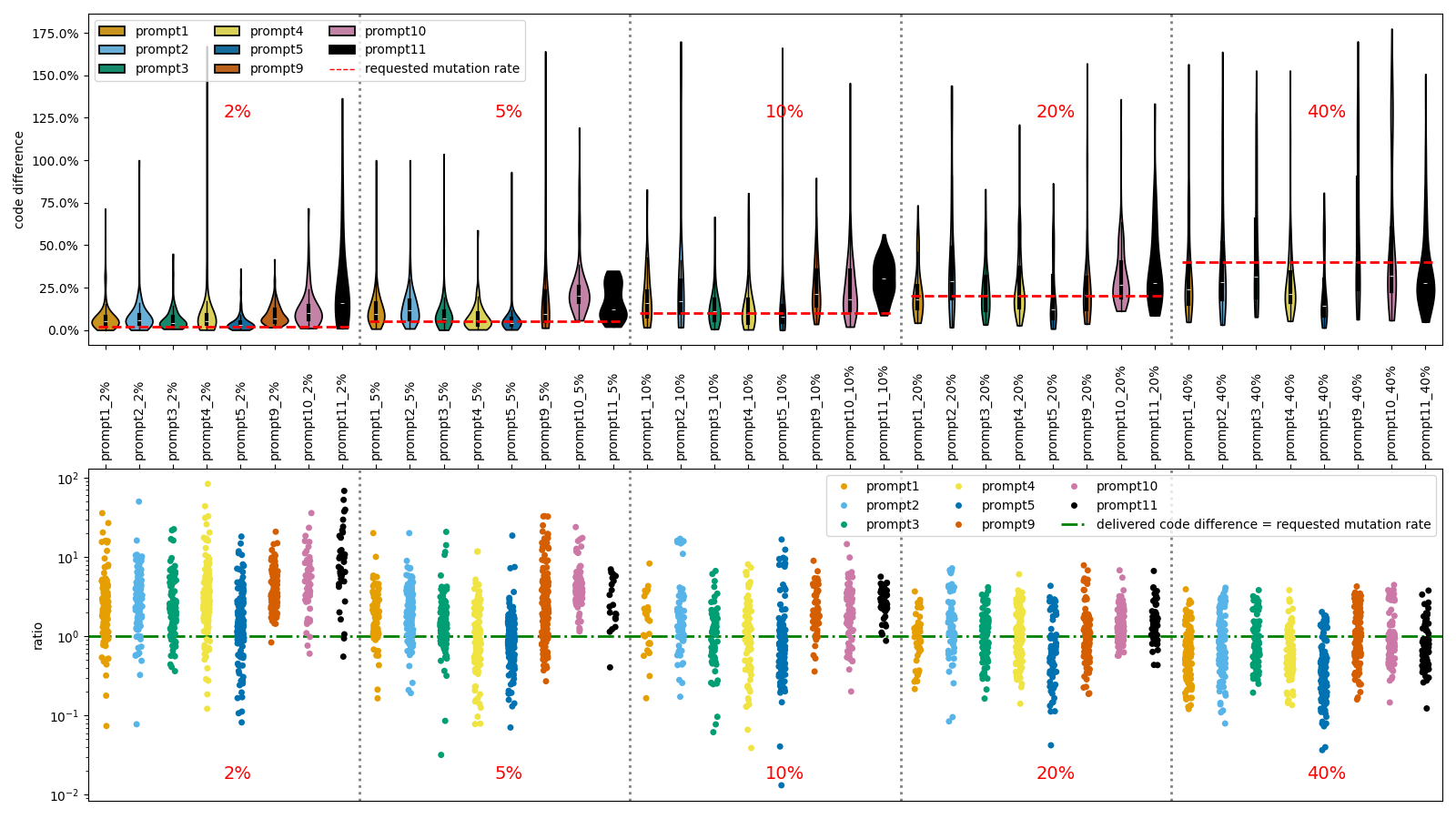} 
    \caption{GPT-4o}
    \label{fig:aggregated-gpt-4o}
  \end{subfigure}
  \caption{The distribution of code difference of the codes generated with different prompts and requested mutation rates, and scatterplot of the ratio of actual delivered code difference to requested mutation rate for LLaMEA. Results of different mutation rates are separated by gray dotted lines and marked by red text. The red dashed line represents the requested mutation rate, while the green dotted-dashed line shows that the difference in the delivered code for points on this line equals the requested mutation rate. Each data column contains mutated codes from 100 codes generated over 3 runs.
  }
  \label{fig:aggregated}
\end{figure}

%% file: convergence.tex
\begin{figure}[!t]
\centering
  \begin{subfigure}[b]{0.48\textwidth} 
    \includegraphics[trim=10 10 10 10,clip,width=\textwidth]{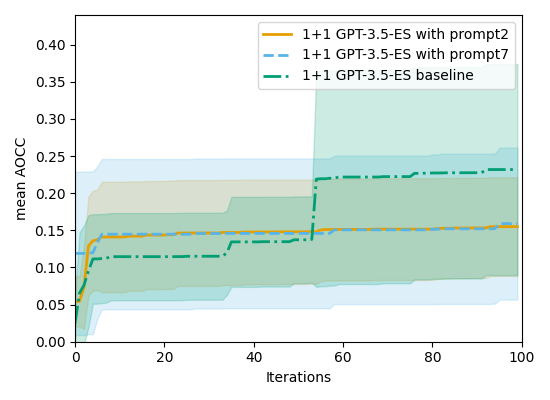}
    \caption{GPT-3.5-turbo}
    \label{fig:convergence-gpt-3.5-turbo}
  \end{subfigure}
  \begin{subfigure}[b]{0.48\textwidth}
    \includegraphics[trim=10 10 10 10,clip,width=\textwidth]{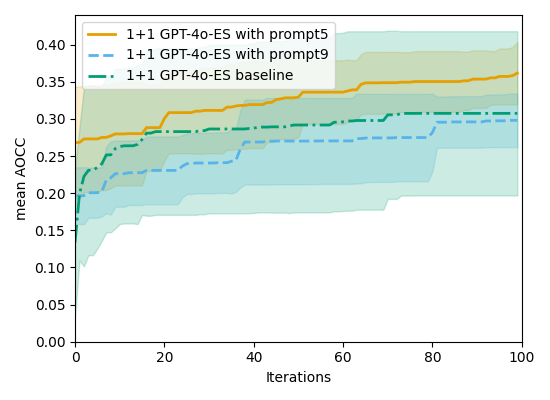}
    \caption{GPT-4o}
    \label{fig:convergence-gpt-4o}
  \end{subfigure}
    \caption{Mean convergence curves (best-so-far algorithm scores) over the 5 different runs for each selected prompt and LLM. Baselines are the raw data from~\cite{van2024llamea}. Shaded areas denote the standard deviation of the best-so-far.
    }
    \label{fig:convergence}
\end{figure}

%% file: code_diff.tex
\begin{figure}[!t]
\centering
  \begin{subfigure}[b]{0.48\textwidth} 
    \includegraphics[trim=10 10 10 10,clip,width=\textwidth]{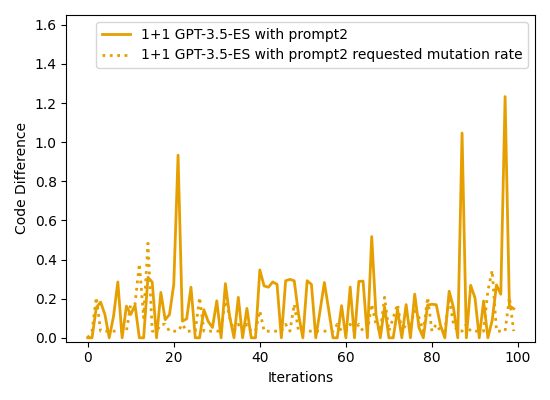}
    \caption{GPT-3.5-turbo with prompt2}
    \label{fig:code_diff-gpt-3.5-turbo-exp1}
  \end{subfigure}
  \begin{subfigure}[b]{0.48\textwidth}
    \includegraphics[trim=10 10 10 10,clip,width=\textwidth]{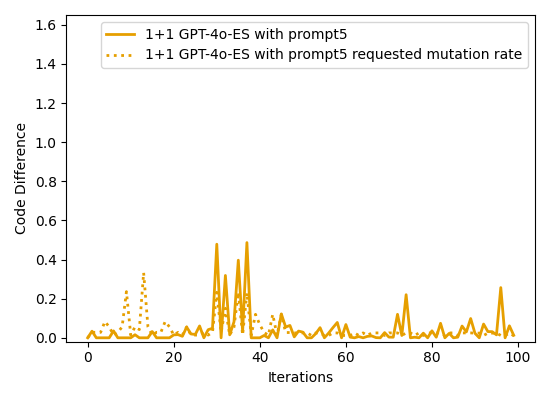}
    \caption{GPT-4o with prompt5}
    \label{fig:code_diff-gpt-4o-exp1}
  \end{subfigure}
  \begin{subfigure}[b]{0.48\textwidth} 
    \includegraphics[trim=10 10 10 10,clip,width=\textwidth]{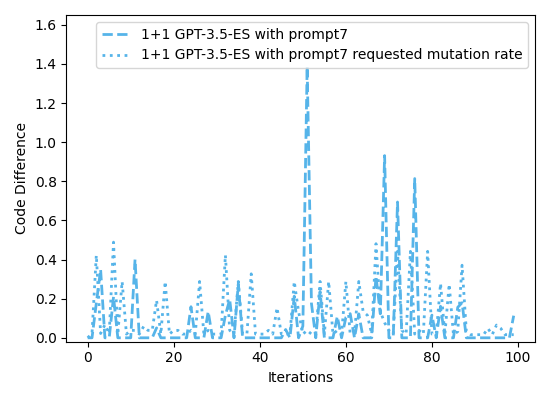}
    \caption{GPT-3.5-turbo with prompt7}
    \label{fig:code_diff-gpt-3.5-turbo-exp2}
  \end{subfigure}
  \begin{subfigure}[b]{0.48\textwidth}
    \includegraphics[trim=10 10 10 10,clip,width=\textwidth]{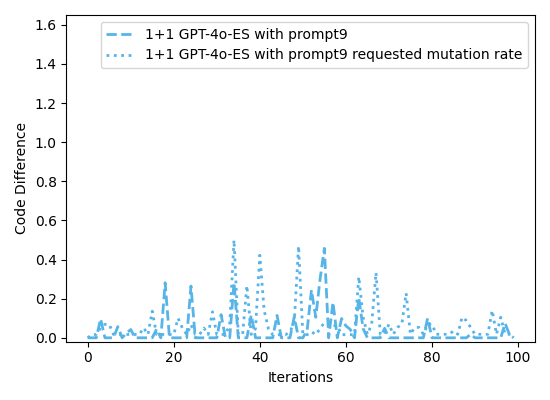}
    \caption{GPT-4o with prompt9}
    \label{fig:code_diff-gpt-4o-exp2}
  \end{subfigure}
  \begin{subfigure}[b]{0.48\textwidth} 
    \includegraphics[trim=10 10 10 10,clip,width=\textwidth]{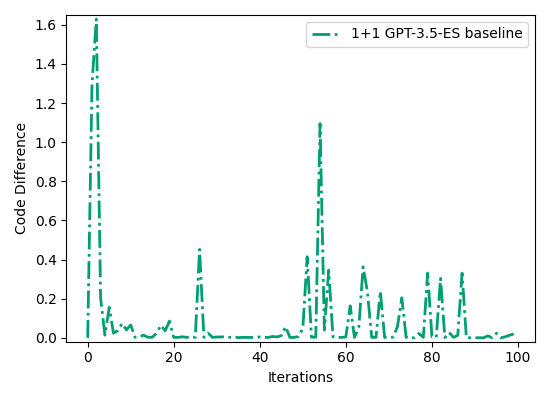}
    \caption{GPT-3.5-turbo baseline}
    \label{fig:code_diff-gpt-3.5-turbo-baseline}
  \end{subfigure}
  \begin{subfigure}[b]{0.48\textwidth}
    \includegraphics[trim=10 10 10 10,clip,width=\textwidth]{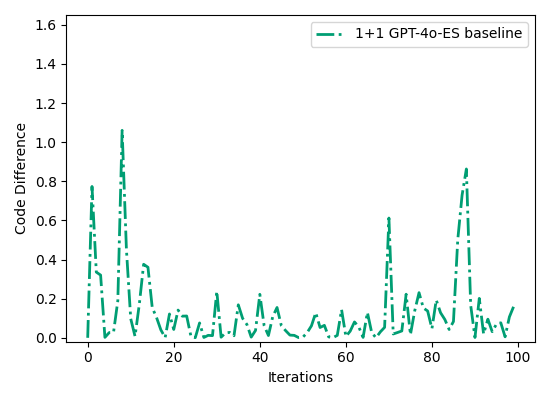}
    \caption{GPT-4o baseline}
    \label{fig:code_diff-gpt-4o-baseline}
  \end{subfigure}
    \caption{Example code difference of 1 run out of 5 runs for each selected prompt and LLM. Baselines are the raw data from~\cite{van2024llamea}. For Figs.~\ref{fig:code_diff-gpt-3.5-turbo-exp1}, \ref{fig:code_diff-gpt-4o-exp1}, \ref{fig:code_diff-gpt-3.5-turbo-exp2}, and \ref{fig:code_diff-gpt-4o-exp2}, the dotted lines represent requested mutation rate. 
    }
    \label{fig:code_diff}
\end{figure}